\documentclass[final]{cvpr}

\usepackage{times}
\usepackage{epsfig}
\usepackage{graphics}
\usepackage{graphicx}
\usepackage{amsmath}
\usepackage{amsthm}
\usepackage{amssymb}
\newtheorem{prop}{Proposition}
\usepackage{mathtools}
\usepackage[linesnumbered,ruled,vlined]{algorithm2e}
\usepackage{xcolor}
\usepackage{multirow}
\usepackage{multicol}
\usepackage{hhline}
\usepackage{booktabs}
\usepackage{tabularx}
\usepackage{anyfontsize}
\usepackage{threeparttable}
\usepackage{ragged2e}
\usepackage{makecell}

\usepackage[pagebackref=true,breaklinks=true,colorlinks,bookmarks=false]{hyperref}

\def\confYear{CVPR 2021}

\begin{document}

\title{Style Intervention: How to Achieve Spatial Disentanglement with \\ Style-based Generators?}

\author{Yunfan Liu$^{1,3}$\qquad Qi Li$^{1,2}$\qquad Zhenan Sun$^{1,2,3}$\qquad Tieniu Tan$^{1,2,3}$\\
$^{1}$ Center for Research on Intelligent Perception and Computing, CASIA\\
$^{2}$ National Laboratory of Pattern Recognition, CASIA\\
$^{3}$ School of Artificial Intelligence, University of Chinese Academy of Sciences\\
{\tt\small yunfan.liu@cripac.ia.ac.cn, \{qli, znsun, tnt\}@nlpr.ia.ac.cn}
}

\maketitle

\begin{abstract}
Generative Adversarial Networks (GANs) with style-based generators (\eg StyleGAN) successfully enable semantic control over image synthesis, and recent studies have also revealed that interpretable image translations could be obtained by modifying the latent code.
However, in terms of the low-level image content, traveling in the latent space would lead to `spatially entangled changes' in corresponding images, which is undesirable in many real-world applications where local editing is required.
To solve this problem, we analyze properties of the \textbf{`style space'} and explore the possibility of controlling the local translation with pre-trained style-based generators.
Concretely, we propose \textbf{`Style Intervention'}, a lightweight optimization-based algorithm which could adapt to arbitrary input images and render natural translation effects under flexible objectives.
We verify the performance of the proposed framework in facial attribute editing on high-resolution images, where both photo-realism and consistency are required.
Extensive qualitative results demonstrate the effectiveness of our method, and quantitative measurements also show that the proposed algorithm outperforms state-of-the-art benchmarks in various aspects.
\end{abstract}

\section{Introduction}\label{sec:introdution}
Generative Adversarial Networks (GANs)~\cite{goodfellow2014generative} with style-based generators have received significant research attention as they enable interpretable control of synthesized images at different scales~\cite{karras2019style}. 
The main difference between these networks and traditional generators is that each convolutional layer is weighted via adaptive instance normalization (AdaIN~\cite{huang2017adain,park2019SPADE}) with parameters computed based on the input latent code.
Therefore, adjusting these~\textit{`style coefficients'} modifies the relative strength of feature maps at each scale, which enables direct manipulation of the generation result and thus is widely adopted in many studies~\cite{jiang2020psgan,men2020controllable,wang2020neural,Zhu_2020_CVPR}.

\begin{figure}[t]
\begin{center}
\includegraphics[width=1.0\linewidth]{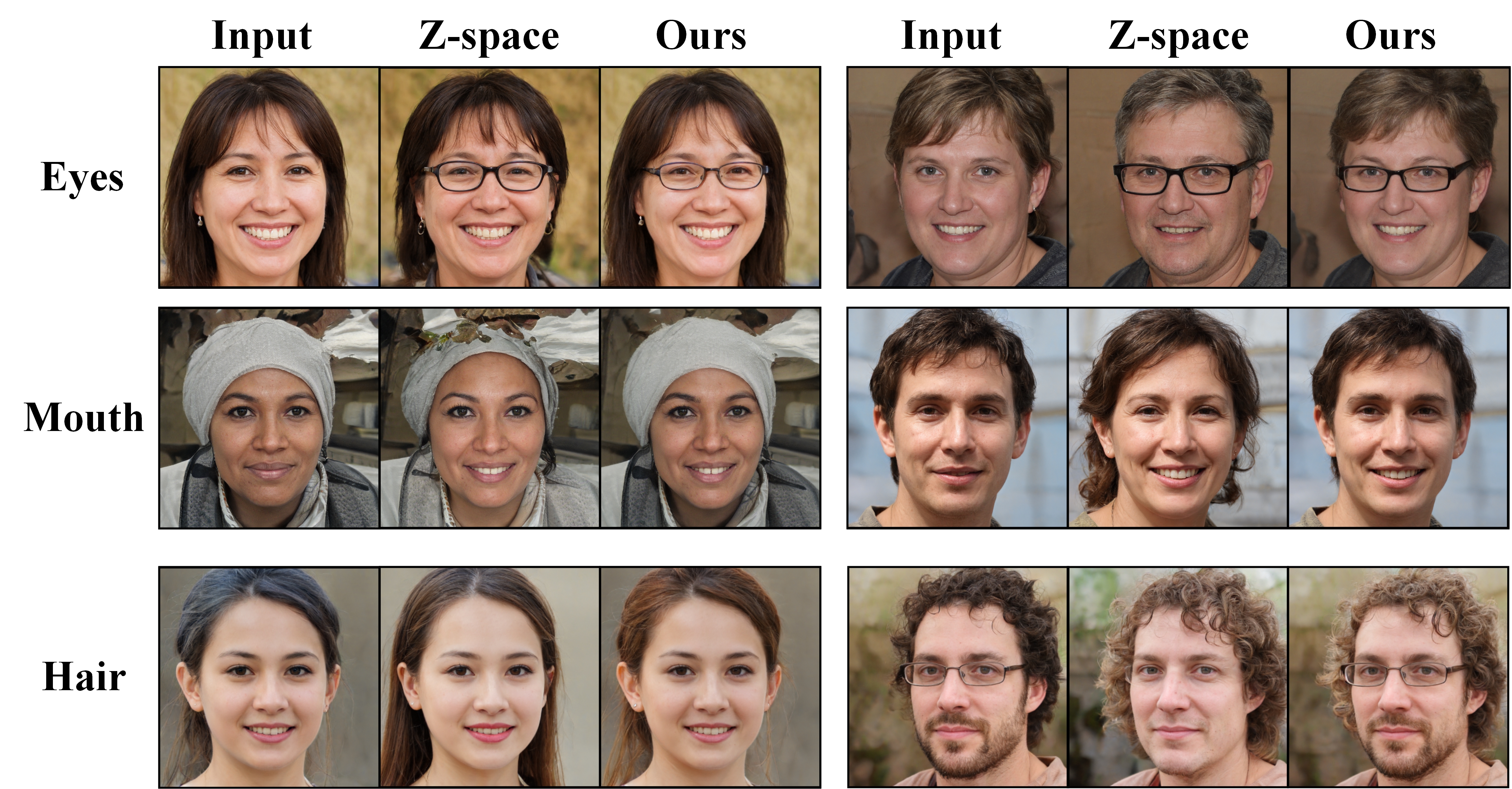}
\end{center}
\caption{Examples of spatial entanglement in image translation. Names of the object to be manipulated are labeled on the left and the attribute for each row is annotated underneath. 
}
\label{fig:teaser}
\end{figure}

Due to the high controllability of style-based generators, numerous studies have been conducted to analyze the organization of latent space and solve for latent directions for interpretable semantic changes.
Unsupervised approaches~\cite{goetschalckx2019ganalyze,harkonen2020ganspace,shen2020closed,voynov2020unsupervised} use classical unsupervised machine learning techniques, \eg Principal Component Analysis (PCA), to discover the collection of latent directions for semantic control.
Supervised methods~\cite{abdal2020styleflow,bau2020semantic,shen2020interpreting,tewari2020stylerig}, on the other hand, edit latent codes under the guidance of attribute labels in certain translation tasks.
However, although semantic attributes of the input image could be correctly manipulated by existing methods, the translation of specific image content remains `spatially entangled', as shown in Figure~\ref{fig:teaser}. 
This makes existing methods impractical  in many real-world applications where local editing is required (\eg portrait manipulation).

One intuitive solution for this problem would be introducing a binary mask to select the foreground and merge background to the edited image.
However, it is extremely difficult, if not impossible, to obtain an accurate mask for translation with large shape deformations.
From another perspective, manipulating feature maps along the spatial dimension~\cite{alharbi2020disentangled,suzuki2018spatially},~\eg introducing `dense style code' which is spatially-variant, would help restrict image changes within the target area.
However, this would heavily increase the number of parameters in the model as well as the cost of training, especially for images with high resolutions.

Notably, given any object-level visual concept (\eg a facial component) in an image, both its semantic and spatial information are encoded in certain internal representations of the generator~\cite{netdissect2017,bau2019gandissect}.
Therefore, limiting manipulations to normalizing coefficients (\ie style codes) of those feature maps would largely reduce the influence on synthesizing other image content.
To this end, instead of the highly entangled latent space, we investigate the \textbf{`style space'}, \ie the space spanned by all possible style vectors, to achieve fine-grained controls on local translations.
Concretely, we propose a lightweight optimization-based framework, named \textbf{`Style Intervention'}, that could discover feature maps closely relevant to the target visual concept and perform precise image manipulation.
Unlike previous methods, Style Intervention does not train any additional network (as in~\cite{bau2020semantic,tewari2020stylerig}), or require the annotation of any extra attribute (as in~\cite{shen2020interpreting}), since it makes the full use of intermediate variables.
Although we stick to the generator of StyleGANv2~\cite{karras2020analyzing} in this paper, Style Intervention is a general framework which could adapt to any style-based generator, and the objective function could also be flexibly customized to suit various translation tasks.
We verify the effectiveness of the proposed algorithm by the problem of facial attribute editing on high-resolution images, and extensive experimental results demonstrate that our method outperforms previous state-of-the-art in terms of both visual fidelity and spatial disentanglement.

Our main contributions could be summarized as follows,
\begin{enumerate}
   \item Unlike most previous studies focusing on the input latent space, we investigate the \textbf{style space} of style-based generators to achieve spatial disentanglement. To the best of our knowledge, this work is the first to explore the possibility of controlling precise \textit{local} translations with \textit{globally} applied style coefficients.

   \item We propose~\textbf{Style Intervention}, a general optimization-based solution for precise image translation applicable to any style-based generator. It is lightweight, as no addition network structure or annotation data is required, and flexible, as the objective function could be customized to fit any translation task.

   \item We choose the task of facial attribute editing as the test bench of the propose algorithm. Extensive experiments are conducted on high-resolution images and results demonstrate the advantage of our method in controlling local translation while ensuring the visual fidelity of generated images.
\end{enumerate}

\section{Related Work}

\subsection{Latent Space Analysis of GANs}
Due to the remarkable success of StyleGAN~\cite{karras2019style} and StyleGANv2~\cite{karras2020analyzing} in image synthesis, many studies have been conducted to analyze the property of their latent space where style codes are computed.

Unsupervised approaches~\cite{goetschalckx2019ganalyze,harkonen2020ganspace,shen2020closed,voynov2020unsupervised} adopt classical unsupervised machine learning techniques, \eg PCA, to solve for important latent directions and interpret their semantic meanings in generation results. 
Supervised methods, on the other hand, solve for the manipulated latent vector with the supervision of semantic labels.
StyleRig~\cite{tewari2020stylerig} further parameterizes the latent space of StyleGAN using a 3DMM model and achieve accurate 3D control for portrait images.
GANPaint~\cite{bau2020semantic} proposes to use another deep network to generate perturbations that correct intermediate feature maps of the generator.
InterFaceGAN~\cite{shen2020interpreting} trains linear support vector machines (SVMs) to classify latent codes based on attribute annotations, where the normal vector of each hyperplane represents the latent direction of the corresponding attribute.

\subsection{Facial Attribute Editing with GANs}
Facial attribute editing (FAE) is one of the most important research topics of image translation due to its wide range of practical applications.
FAE aims at authentically manipulating the target attribute of an input face image while keep irrelevant image content intact.
To solve this problem, numerous methods built on conditional GAN models~\cite{choi2018stargan,he2019attgan,lee2020maskgan,liu2019stgan,Wu_2019_ICCV} are proposed, and they typically focus on designing new loss functions or improving network structures.
However, when it comes to images with high resolutions (\eg $1024\times 1024$), training these models from scratch is computationally expensive and the quality of generation results is hardly satisfying.
Therefore, researchers have resorted to reusing large-scale style-based generators pre-trained on high-quality images as powerful generative priors, and control their behavior by manipulating the latent code along interpretable semantic directions.


\section{Method}
A style-based generator $G:\mathcal{Z}\rightarrow \mathcal{I}$ consists of a mapping network $f:\mathcal{Z}\rightarrow \mathcal{S}$, which computes the style code $\mathbf{s}$ based on the input latent vector $\mathbf{z}$, followed by a synthesis network $h:\mathcal{S}\rightarrow \mathcal{I}$, which modulates each feature map by component of $\mathbf{s}$ and renders the final output ($G=f\circ h$).
Due to such complex structure of $G$, traveling in the input latent space $\mathcal{Z}$, as in most previous studies~\cite{bau2020semantic,goetschalckx2019ganalyze,harkonen2020ganspace,shen2020interpreting,shen2020closed,tewari2020stylerig,voynov2020unsupervised}, would inevitably cause spatial entanglement.
This is because editing one single component of $\mathbf{z}$ would affect style codes $\mathbf{s}=f(\mathbf{z})$ of superfluous feature maps, which increases the chance of modifying non-target image content. 
Therefore, the fundamental solution for disentangling translations of the target visual concept $c$ in the spatial dimension, is to restrict the modification of $\mathbf{s}$ to those channels which are responsible for the synthesis of $c$ while irrelevant to other objects.

\begin{figure}[t]
\begin{center}
\includegraphics[width=1.0\linewidth]{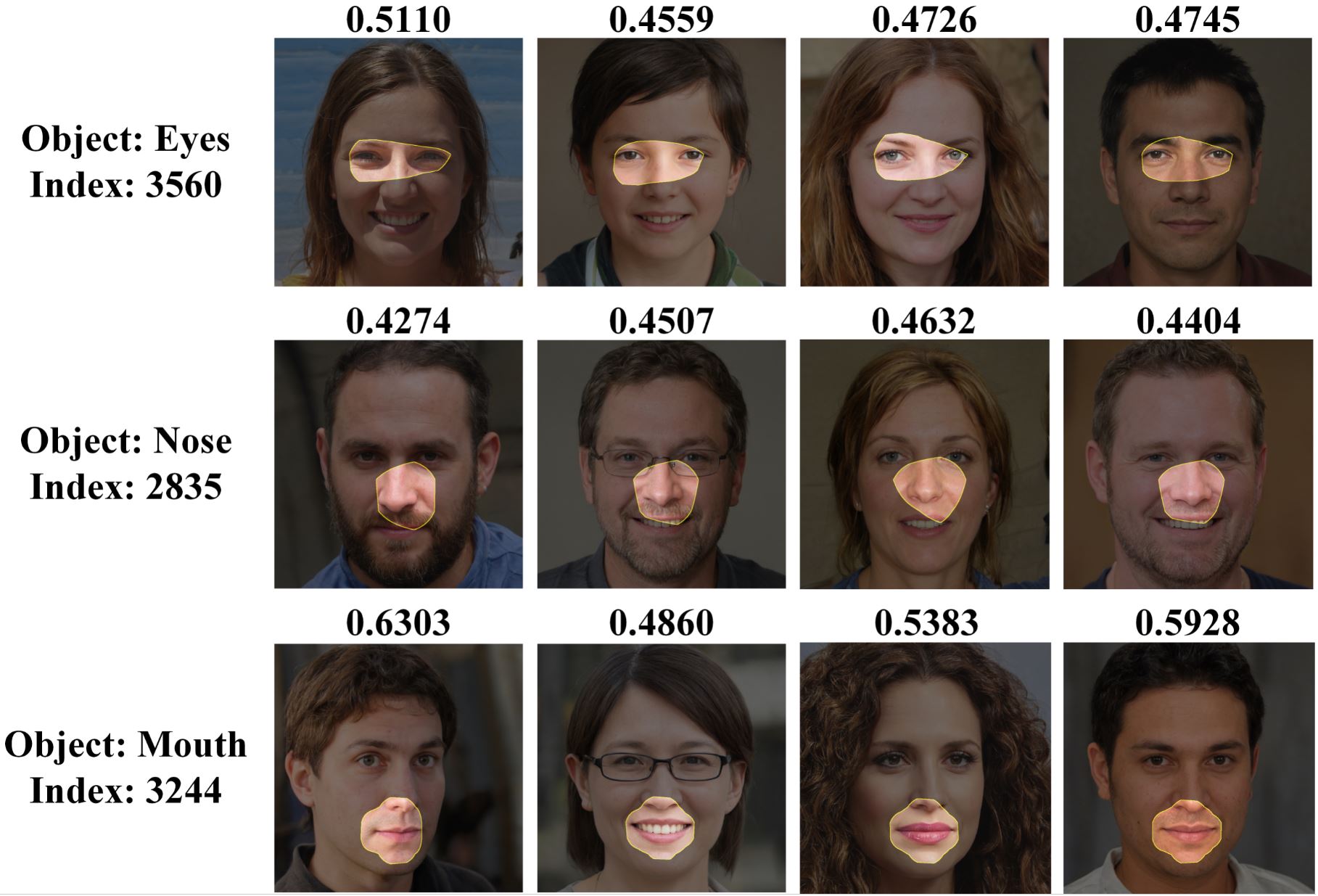}
\end{center}
\caption{Visualization of feature maps at the same level of StyleGANv2~\cite{karras2020analyzing} pre-trained on FFHQ~\cite{karras2019style}. We compute the binarized feature map by setting the top 5\% activated spatial locations to 1 and others to 0. The intersection-over-union (IoU) value between the binarized feature map and the segmentation result of $c$ for each sample is labeled above the image.}
\label{fig:iou}
\end{figure}

Similar to previous studies on model intervention~\cite{netdissect2017,bau2019gandissect,simonyan2013deep}, we observe that feature maps at certain channels consistently align with the spatial distribution of various classes of $c$, after being upsampled and thresholded (see Figure~\ref{fig:iou}).
This demonstrates the potential feasibility of localizing image translations by modifying style codes although they are globally applied.
Following the convention in previous studies~\cite{bau2019gandissect,netdissect2017}, we denote the collection of all feature maps as $\mathbb{U}$
, and we use the subscript $c$ and $\overline{c}$ to distinguish components (\eg $\mathbb{U}$ and $\mathbf{s}$) responsible for generating $c$ from irrelevant ones.

\subsection{Translation in the Style Space $\mathcal{S}$}\label{sec:translation_style_space}
Following the previous discussion, two immediate questions one might ask would be 1)~\textit{how to solve the division $\mathbb{U}=(\mathbb{U}_c, \mathbb{U}_{\overline{c}})$}, and 2)~\textit{afterwards, how to edit $\mathbf{s}$ to precisely manipulate the semantic of $c$}. 
These questions are extremely difficult to answer due to the intrinsic complexity of style-based generators, as modifying one single style code would affect all subsequent feature maps and thus may cause a series of unpredictable consequences.

However, this issue could be considered from another perspective, that is, the property of the resultant style code, denoted as $\hat{\mathbf{s}}$, after the manipulation.
Based on \textbf{Proposition}~\ref{prop:ideal_style_code}, if $f_{\alpha}$ is implemented by a separating hyperplane with the normal vector denoted as $n_{\alpha}$, the corresponding displacement vector $\Delta \hat{\mathbf{s}}=\hat{\mathbf{s}}-\mathbf{s}$ should satisfy the following conditions:
\begin{equation}\label{eq:parallel}
\Delta \hat{\mathbf{s}}\approx k\cdot n_{\alpha}, \text{where}~k~\text{is constant}
\end{equation}
\begin{equation}\label{eq:equals_to_zero}
\Delta \hat{\mathbf{s}}_{\overline{c}}=0
\end{equation}
Eq.~\ref{eq:parallel} indicates that $\hat{\mathbf{s}}$ should render the desired translation effect of $\alpha$, while Eq.~\ref{eq:equals_to_zero} suggests that $\mathbf{s}_{\overline{c}}$ is supposed to remain intact.
In this way, interpolating along $\Delta \hat{\mathbf{s}}$ would lead to the desired edited image after forwarding through $h$.


How to approach $\Delta \hat{\mathbf{s}}$?
Based on Eq.~\ref{eq:parallel} and Eq.~\ref{eq:equals_to_zero}, it is natural to approximate it using the \textbf{normal vector of a separating hyperplane in $\mathcal{S}$}, denoted as $\Delta \mathbf{s}_n$, which classifies style codes according to labels of the translated attribute $\alpha$.
Moreover, \textbf{sparsity regularization} is imposed on $\Delta \mathbf{s}_n$ to meet the requirement of Eq.~\ref{eq:equals_to_zero}, which encourages the displacement vector to modify fewer style codes.
In this way, we \textbf{unify} translating the input image and reducing irrelevant changes in a single framework.

\begin{figure}[t]
\begin{center}
\includegraphics[width=0.9\linewidth]{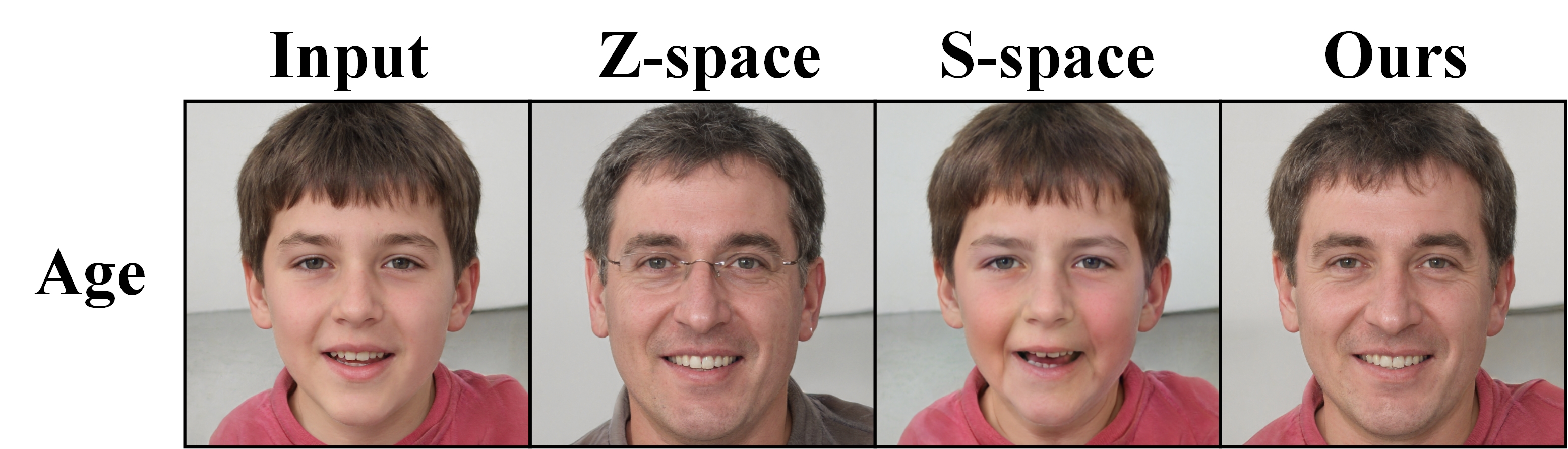}
\end{center}
\caption{An example of interpolating in the latent space $\mathcal{Z}$ and $\mathcal{S}$. Our method is able to combine the advantage of editing in both spaces and could generate realistic output images.}
\label{fig:w_plus_interp}
\end{figure}

\begin{prop}\label{prop:ideal_style_code}
Given the input style code $\mathbf{s}$, a style-based synthesis network $h:\mathcal{S}\rightarrow \mathcal{I}$, and an attribute classifier $f_\alpha:\mathcal{S}\rightarrow \mathbb{R}$ for the attribute $\alpha$ of $c$, where
\begin{align}
f_\alpha(\mathbf{s}) \left\{ \begin{array}{ll}
>0 & \text{for}~h(\mathbf{s})~\text{with positive label of}~\alpha \\
<0 & \text{for}~h(\mathbf{s})~\text{with negative label of}~\alpha
\end{array}\right.
\end{align}
the style code $\hat{\mathbf{s}}$ of an ideally translated image should satisfy the following conditions,
\begin{equation}
f_\alpha(\mathbf{s})\cdot f_\alpha(\mathbf{\hat{s}}) < 0
\end{equation}
\begin{equation}\label{eq:style_equality}
\hat{\mathbf{s}}_{\overline{c}}=\mathbf{s}_{\overline{c}}
\end{equation}

\end{prop}

\subsection{Style Intervention with Self-supervision}
Although $\Delta \mathbf{s}_n$ has properties similar to the optimal style displacement vector $\Delta \hat{\mathbf{s}}$, it does not provide adequate clues to render natural translation effects.
This is because $\Delta \mathbf{s}_n$ is obtained via a discriminative task, and thus only contains the minimum information to translate the input image so as to change the attribute label of $c$, which does not guarantee the realism of generation results (as shown in Figure~\ref{fig:w_plus_interp}).

However, since the result of translating in $\mathcal{Z}$ ($\mathbf{z}\rightarrow \mathbf{z'}$) is visually plausible, the corresponding displacement vector in $\mathcal{S}$, \ie $\Delta \mathbf{s}_z= f(\mathbf{z}')-f(\mathbf{z})$, should contain sufficient information to synthesize rich textural details.
To this end, we naturally propose to combine the translation in both $\mathcal{Z}$, which introduces realistic semantic changes but is short on spatial disentanglement, and $\mathcal{S}$, which is able to exert precise modifications but may be less satisfactory in generating photo-realistic results, for authentic and accurate manipulation of image components.





To achieve this, an intuitive method would be changing each component in $\Delta \textbf{s}_z$ according to the corresponding value in $\Delta \textbf{s}_n$, and check whether it enhances the translation result.
However, in most cases, the generation of $c$ jointly depends on multiple feature maps, and thus solving for one single component in $\Delta \textbf{s}$ at a time could not lead to reasonable results.
Therefore, we propose to tackle the problem by solving for the intervention coefficient $\Lambda=\{\lambda_i\}_{i=1}^n$, where $n$ is the number of convolutional layers.
Each $\lambda_i\in [0,1]^{l_i}$ indicates the degree of intervention for all channels in the $i$~th convolutional layer, where ${l_i}$ is the the number of feature maps within.
Hence, the merged displacement of style code, denoted as $\Delta \mathbf{s}_m$, could be computed as
\begin{equation}\label{eq:combine}
\Delta \mathbf{s}_m(\Lambda) = (\mathbf{1}-\Lambda)\cdot \Delta \textbf{s}_z + \Lambda\cdot \Delta \textbf{s}_n
\end{equation}

The self-supervising objective function for solve the optimal intervention coefficient $\Lambda^*$ contains three parts:
\begin{itemize}
   \item \textbf{Pixel-level Loss $\mathcal{L}_{pix}$ :} We explicitly penalize $\Lambda$ for modifying image content other than $c$. 
   Let us denote the binary mask for $c$ obtained from the semantic segmentation of $G(z)$ as $m_c$, where pixel values within $c$ is set to 1 and elsewhere to 0.
   The pixel-level loss could be formulated as
   \begin{equation}\label{eq:L_pix}
        \mathcal{L}_{pix}=\|(1-m_c)\cdot (h(\mathbf{s}+\Delta \mathbf{s}_m(\Lambda))-h(\mathbf{s}))\|_2
   \end{equation}

   \item \textbf{Attribute Loss $\mathcal{L}_{attr}$ :} To ensure that the edited image does present the target semantic change, we introduce the attribute loss $\mathcal{L}_{attr}$, which could be computed as
   \begin{equation}
        \mathcal{L}_{attr}=-\frac{\Delta \mathbf{s}_n\cdot \Delta \mathbf{s}_m(\Lambda)}{\|\Delta \mathbf{s}_n\|_2\cdot \|\Delta \mathbf{s}_m(\Lambda)\|_2}
   \end{equation}
   It is obvious that $\mathcal{L}_{attr}$ is exactly the cosine similarity between $\Delta \mathbf{s}_n$ and $\Delta \mathbf{s}_m(\Lambda)$, which encourage the resultant style code $\Delta \mathbf{s}_m$ to still render the desired attribute change presented by $\Delta \mathbf{s}_n$.

   \item \textbf{L2-norm Loss $\mathcal{L}_{norm}$ :} Although introducing $\Delta \textbf{s}_n$ helps with disentangling spatial changes in image, it inevitably makes the resultant style code deviate from the manifold expanded by $\mathbf{s}=f(\mathbf{z})$, which could possibly make the corresponding image less natural.
   For this reason, we would like to limit the overall degree of intervention using the L2-norm Loss, which could be written as $\mathcal{L}_{norm}(\Lambda)=\|\Lambda\|_2$.
\end{itemize}

Therefore, the overall objective function could be written as 
\begin{equation}
\mathcal{L} = \mathcal{L}_{pix}+\lambda_{attr}\mathcal{L}_{attr}+\lambda_{norm}\mathcal{L}_{norm}
\end{equation}
where $\lambda_{attr}$ and $\lambda_{norm}$ control the relative importance of $\mathcal{L}_{attr}$ and $\mathcal{L}_{norm}$, respectively.
The optimal intervention coefficient $\Lambda^*$ could be solved by $\Lambda^* = arg \min_\Lambda \mathcal{L}$.

Since each convolutional layer is independently normalized~\cite{karras2019style}, we propose a layer-wise optimization algorithm to reduce the interaction between different convolutional layers.
Concretely, from the coarsest level ($i=1$) to the finest level ($i=n$), we solve for one single $\lambda_i$ at a time while keep the intervention coefficient of other layers fixed.
This is because feature maps with low spatial resolutions mainly control global attributes, and thus style codes of these channels should be determined before those adjusting fine-grained details.
Please refer to the supplementary material for details of the proposed algorithm.

\subsection{Differences Compared to Previous Methods}
InterFaceGAN~\cite{shen2020interpreting} is the study closest to ours, but we focus on disentanglement in the spatial dimension instead of the semantic level, which is more practical for real applications.
Moreover, our method does not require labels of extra attributes to perform orthonormalization, not to mention that many image changes, such as identity, could hardly be described by attribute labels.
Image2StyleGAN~\cite{Abdal_2019_ICCV} and its improvement~\cite{Abdal_2020_CVPR} proposes to investigate the extended latent space previous to affine layers, which aims at providing more freedom for GAN inversion.
StyleRig~\cite{tewari2020stylerig} and GANPaint~\cite{bau2020semantic} involve training deep networks along with the StyleGAN model, which are much more computational expensive than our method.
Unsupervised approaches~\cite{goetschalckx2019ganalyze,harkonen2020ganspace,shen2020closed,voynov2020unsupervised} are inappropriate for this task as they focus on discovering interpretable latent semantics, instead of solving for the latent direction for the target attribute.

\begin{figure*}[ht]
\begin{center}
\includegraphics[width=1.0\linewidth]{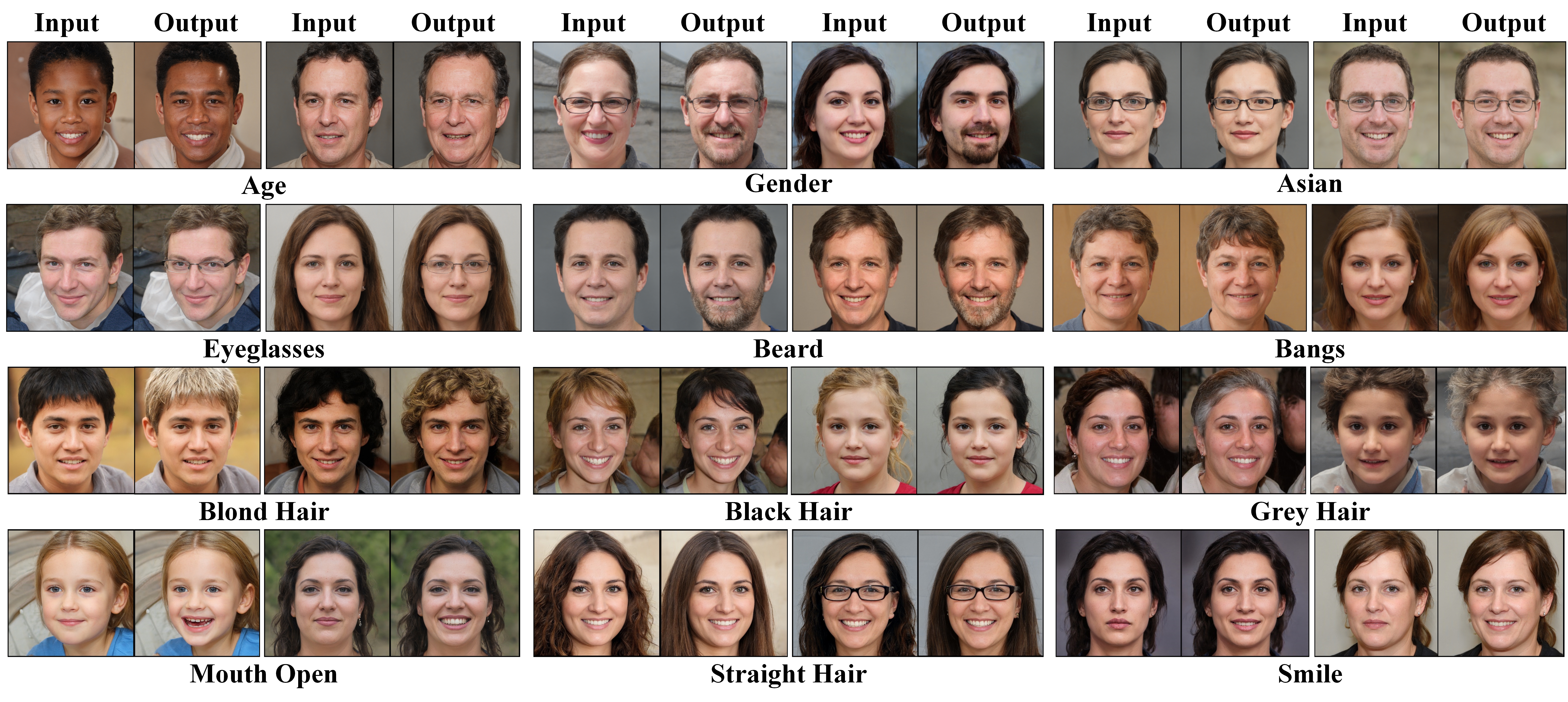}
\end{center}
\caption{Sample results of manipulating different facial component along various semantic directions (labeled on the left).}
\label{fig:single_attribute_editing}
\end{figure*}

\begin{table*}[ht]
\caption{Classification accuracy ($\%$) of linear SVMs on both training and valiadation set. Sparcity values demonstrate the proportion ($\%$) of zero elements in $\Delta \textbf{z}_n$.}
\centering
\resizebox{1.9\columnwidth}{!}{%
\begin{tabular}{ccccccccccccc}
\toprule
Attribute       & Gender  & Age      & Smile    & \makecell{Eye-\\glasses} & \makecell{Mouth\\Open} & Asian  & Bangs   & \makecell{Black\\Hair} & \makecell{Blond\\Hair} & \makecell{Grey\\Hair} & \makecell{Straight\\Hair} & Beard \\
\midrule
Train    & $97.81$ & $100.00$ & $100.00$ & $100.00$ & $100.00$ & $100.00$ & $93.83$ & $96.32$ & $95.48$ & $100.00$  & $86.02$  & $100.00$ \\
Validate & $93.00$ & $98.14$  & $97.70$  & $90.58$  & $97.59$  & $80.40$  & $85.03$ & $82.98$ & $78.54$ & $94.34$   & $81.17$  & $90.14$ \\
Sparcity & $93.60$ & $98.14$  & $96.08$  & $96.92$  & $94.98$  & $96.25$  & $93.61$ & $93.64$ & $93.80$ & $98.55$   & $93.25$  & $97.76$ \\
\bottomrule
\end{tabular}
}
\label{table:facial_component}
\end{table*}

\section{Experiment}
\subsection{Experimental Setup}
\textbf{Data Preparation} We use the facial attribute editing (FAE) task, which has received considerable attention in recent studies on style-based generators, to verify the effectiveness of our method.
In this paper, 12 attributes directions are considered for the qualitative experiment, which is a large extension comparing to previous studies~\cite{harkonen2020ganspace,shen2020interpreting,shen2020closed,voynov2020unsupervised}.
We collect 100,000 synthetic samples and adopt the online face analysis toolkit Face++~\footnote{https://www.faceplusplus.com/} to label 5 facial attributes (`Gender', `Age', `Eyeglasses', `Mouth Open', and `Smile').
For the rest attributes, we randomly select 10,000 samples and manually annotate them, which will be released soon.
Please note our method could be easily extended to other tasks (\eg scene editing), as input images are not required to be faces and other variables (\eg labels and masks) could be easily obtained.

\textbf{Training Configurations} In this paper, we typically focus on the StyleGANv2 model~\cite{karras2020analyzing} pre-trained on the FFHQ dataset~\cite{karras2019style} with resolution $1024\times 1024$.
We concatenate the output of all affine layers to form the style code $\mathbf{s}\in \mathbb{R}^{9088}$.
For the optimization process, we adopt the Adam optimizer and set balancing coefficients $\lambda_{attr}$ and $\lambda_{norm}$ to $1e^{-2}$ and $1e^{-6}$, respectively.
Please refer to the supplementary material for more detailed information of how the mask $m_c$ is computed and what facial components are considered as the target for each attribute.

\subsection{Separability of the Style Space $\mathcal{S}$}
In Section~\ref{sec:translation_style_space}, we approximate the ideal displacement vector in $\mathcal{S}$ by $\Delta \mathbf{s}_n$, \ie the normal vector of a hyperplane classifying style codes by attribute labels.
Therefore, it is necessary to inspect the separability of style codes in the first place since this is the cornerstone of our method.

\begin{figure}[t]
\begin{center}
\includegraphics[width=0.9\linewidth]{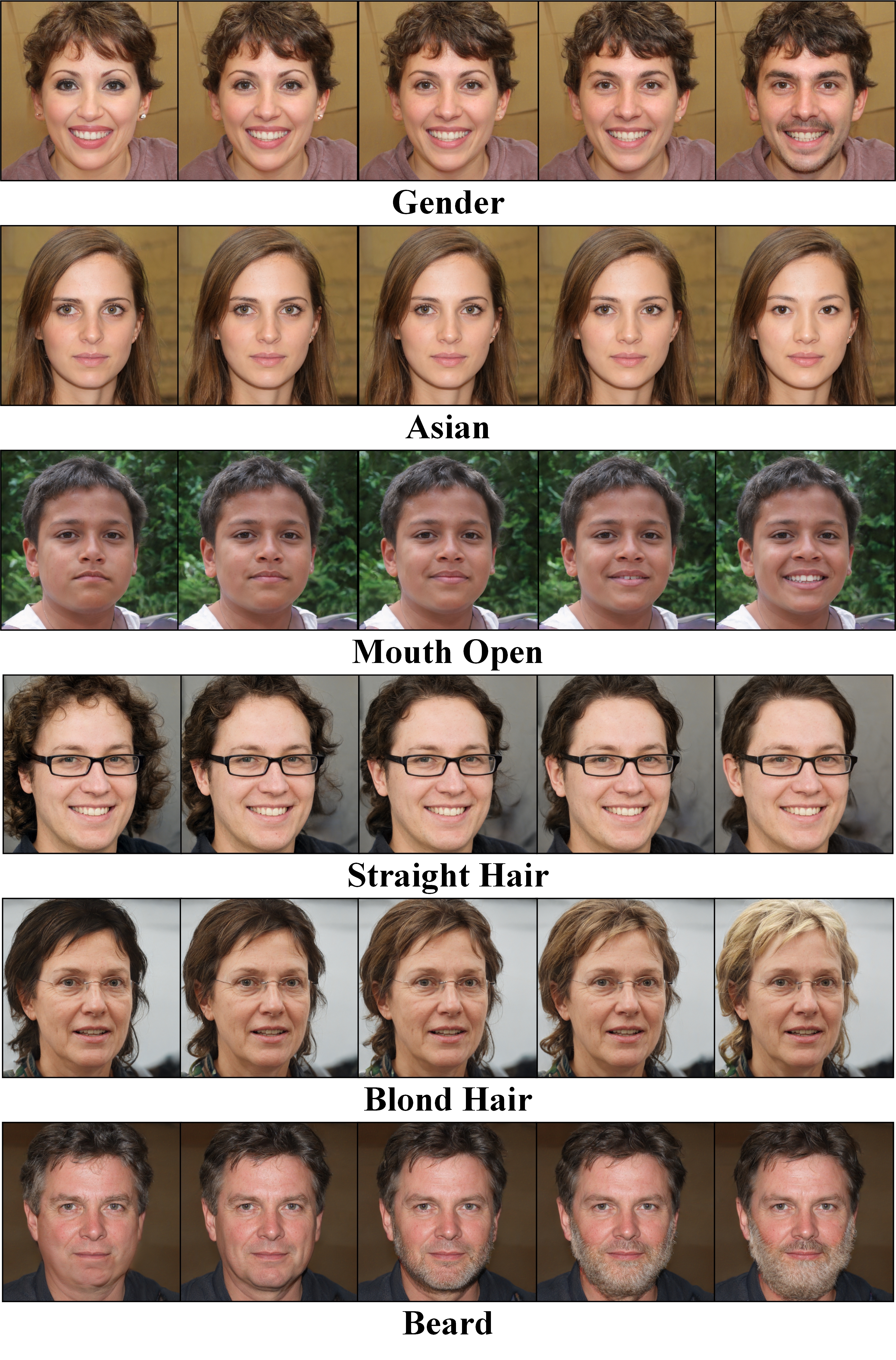}
\end{center}
\caption{Sample results of attribute interpolation (labeled on the left). Note that smooth translation of image attributes is achieved by simply interpolating along $\Delta \mathbf{s}_n$, instead of doing optimization for every single image.}
\label{fig:single_attribute_interpolation}
\end{figure}

According to the result shown in Table~\ref{table:facial_component}, all facial attributes are almost linearly separable in the style space.
This indicates that style codes are not just simple concatenations of normalization coefficients, but also contain discriminative semantic information of image attributes.
Moreover, the high sparsity of normal vectors (over $93\%$) induced by the L1-norm suggests that only a small part of $\mathbb{U}$ is responsible for `flipping' the corresponding attribute, which also proves the possibility for disentangling spatial translation by restricting modifications of the style code to $\mathbf{s}_c$.

\subsection{Spatially Disentangled Image Translation}
From figure~\ref{fig:single_attribute_editing}, it is clear that the proposed method achieves spatial disentanglement on translations of different facial components in various semantic directions.
For example, the hair region could be vividly modified to show different colors or styles, without affecting any other image content.
This is because our method combines interventions in the style space $\mathcal{S}$, which precisely locate feature maps controlling characteristics of the target object, with translations in the latent space $\mathcal{Z}$, which provide rich textural information to render output images.
Notably, our method could also deal with the translation of abstract facial attributes, \ie `Age', `Gender', and `Asian', where we perform the foreground-background disentanglement.

However, one may challenge the effectiveness of our method by attributing the success of precise attribute translations to the explicitly enforced image-level consistency (\ie Eq.~\ref{eq:L_pix}), instead of the disentanglement of style codes.
To this end, we fix $\Lambda^*$ 
and change the coefficient of $\Delta \mathbf{s}_n$ to check if the translation could be interpolated only within the target area.
If results in Figure~\ref{fig:single_attribute_editing} are simply outcomes of the optimization process where unwanted modifications happen to be minimized, this operation would again lead to changes in irrelevant regions.
According to results shown in Figure~\ref{fig:single_attribute_interpolation}, interpolating along $\Delta \mathbf{s}_n$ in the style space does not affect other image content and smooth gradual changes are achieved.
This proves that the result of our method is not simply the solution to a superficial optimization problem, but the outcome of a reasonable combination of translations in two spaces ($\mathcal{Z}$ and $\mathcal{S}$).

\begin{figure}[t]
\begin{center}
\includegraphics[width=1.0\linewidth]{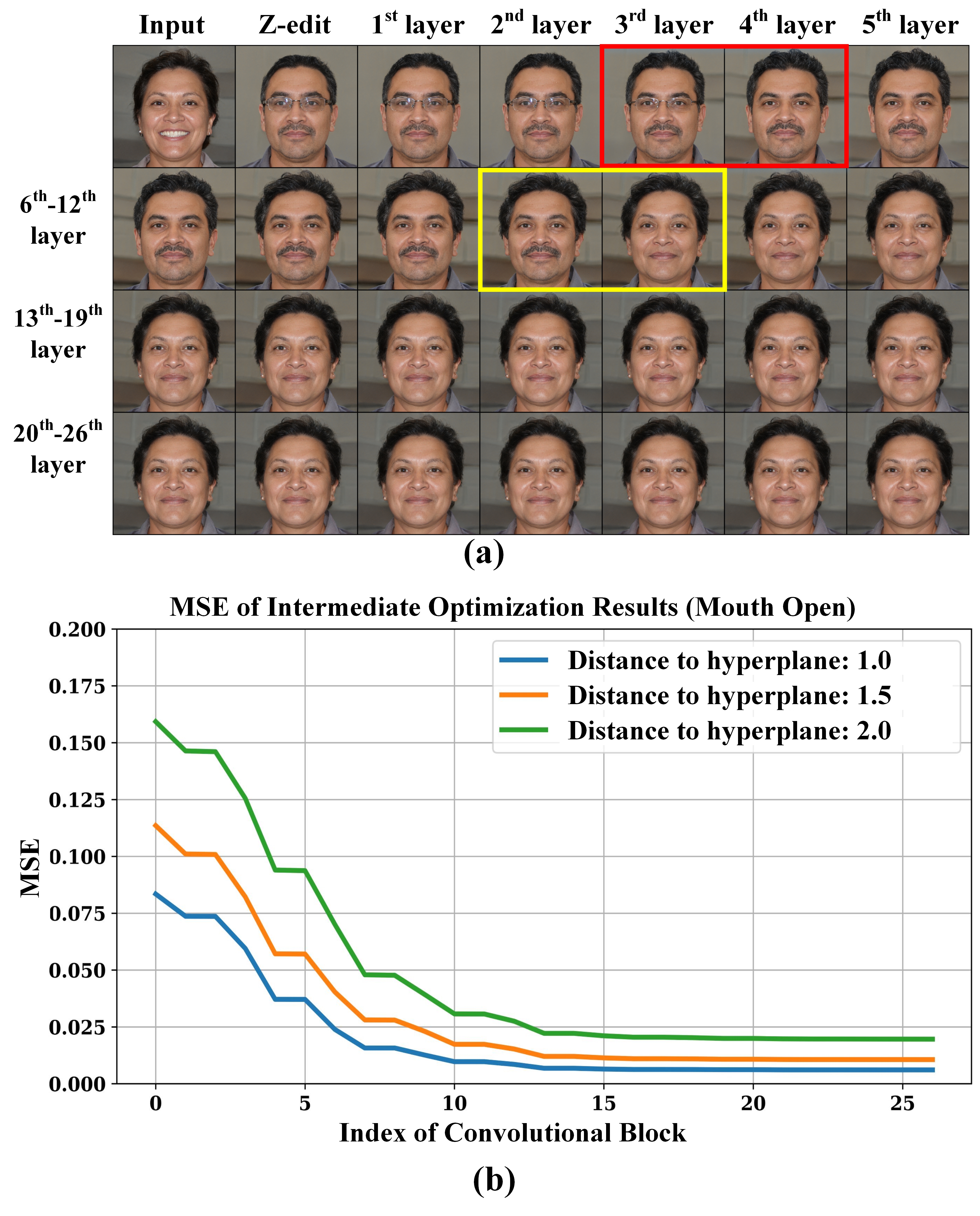}
\end{center}
\caption{Visualization of intermediate results after each optimization step. (a) Resultant images after the optimization of each convolutional layer (there are 26 layers in total); (b) Mean squared loss (MSE) between the input image and the optimization result of each convolutional layer.}
\label{fig:refine_process}
\end{figure}

\subsection{Visualization of Intermediate Results}
Apart from verifying the effectiveness of our method in terms of final generation results, it would also be interesting to investigate the internal mechanism of the proposed algorithm.
To this end, we employ the translation of attribute `Mouth Open' as a case study, and visualize intermediate optimization results to gain a better understanding of how Style Intervention approaches the final output.

According to Figure~\ref{fig:refine_process} (a), it is clear that changes between successive convolutional layers are \textbf{discrete} and \textbf{monotonous}.
The discreteness implies that certain facial component is treated as a whole and embedded in neighboring feature maps, instead of being treated as a collection of local texture patches distributed among all convolutional layers.
For example, eyeglasses incorrectly synthesized by editing in $\mathcal{Z}$ are removed as a whole by the $4^{th}$ layer (red box), rather that gradually disappear through several steps.
Similarly, the mustache is also accurately removed by the $10^{th}$ layer (yellow box).
This demonstrates the possibility of manipulating images on the basis of visual components (\ie with spatial disentanglement) by modifying individual style code, which is the theoretical basis of our method.

As for the monotonicity of image changes, it proves the rationality of performing the optimization in a layer-wise manner, that is, solving for the optimal intervention coefficient of one convolutional layer would not damage the outcome of optimizing previous layers.
This is because the convolutional layers are interdependent since the input of each layer is normalized before being modulated according to the style code.
Therefore, the discrepancy between each intermediate result and the input in non-target image regions keeps decreasing throughout the whole optimization process, which is also verified in Figure~\ref{fig:refine_process} (b) as the error drops monotonously.
We have also tried to enlarge the range of interpolation in $\mathcal{Z}$, and it turns out that although the error between input images and results of editing in $\mathcal{Z}$ increases, the proposed method converges well and reaches low error at the final output layer.

\begin{figure}[t]
\begin{center}
\includegraphics[width=0.83\linewidth]{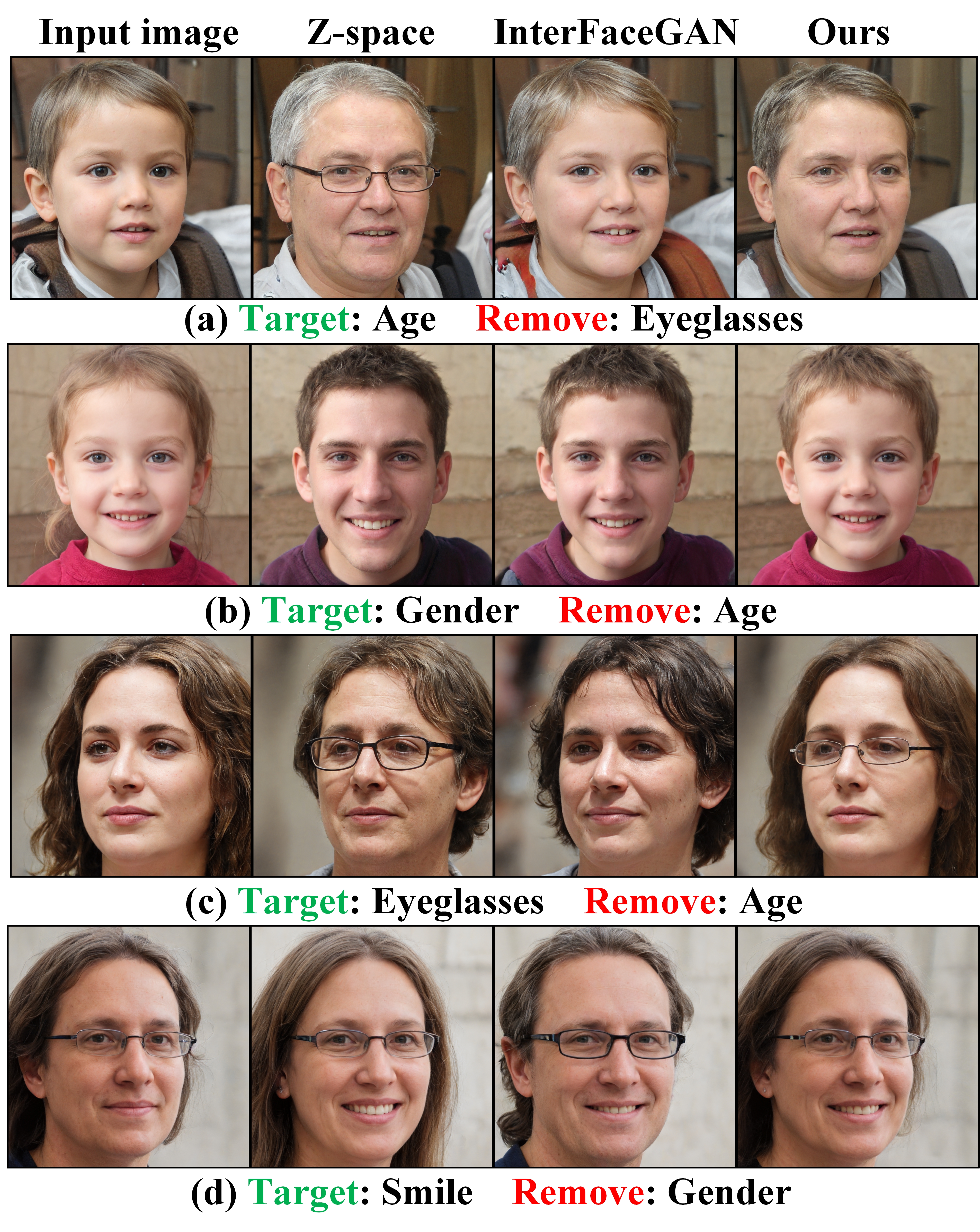}
\end{center}
\caption{Comparison between our method and editing in $\mathcal{Z}$-space, as well as InterFaceGAN~\cite{shen2020interpreting}. Target attributes are marked in green, and unwanted attribute changes are marked in red.}
\label{fig:comparison_z_interfacegan}
\end{figure}

\subsection{Compairson against Benchmark Methods}
\subsubsection{GANs with Style-based Generators}
In order to demonstrate the superiority of the proposed method, we compare manipulation results against several other benchmarks, including editing in $\mathcal{Z}$-space and InterFaceGAN~\cite{shen2020interpreting}.
InterFaceGAN eliminates entangled semantic translations by subtracting the projection on the corresponding direction in the latent space.
To avoid the bias of trained attribute classifiers, we only conduct experiments on four facial attributes that could be recognized by Face++ online APIs for objectivity (we do not include `Mouth Open' as it largely overlaps with `Smile').

\begin{figure}[t]
\begin{center}
\includegraphics[width=0.95\linewidth]{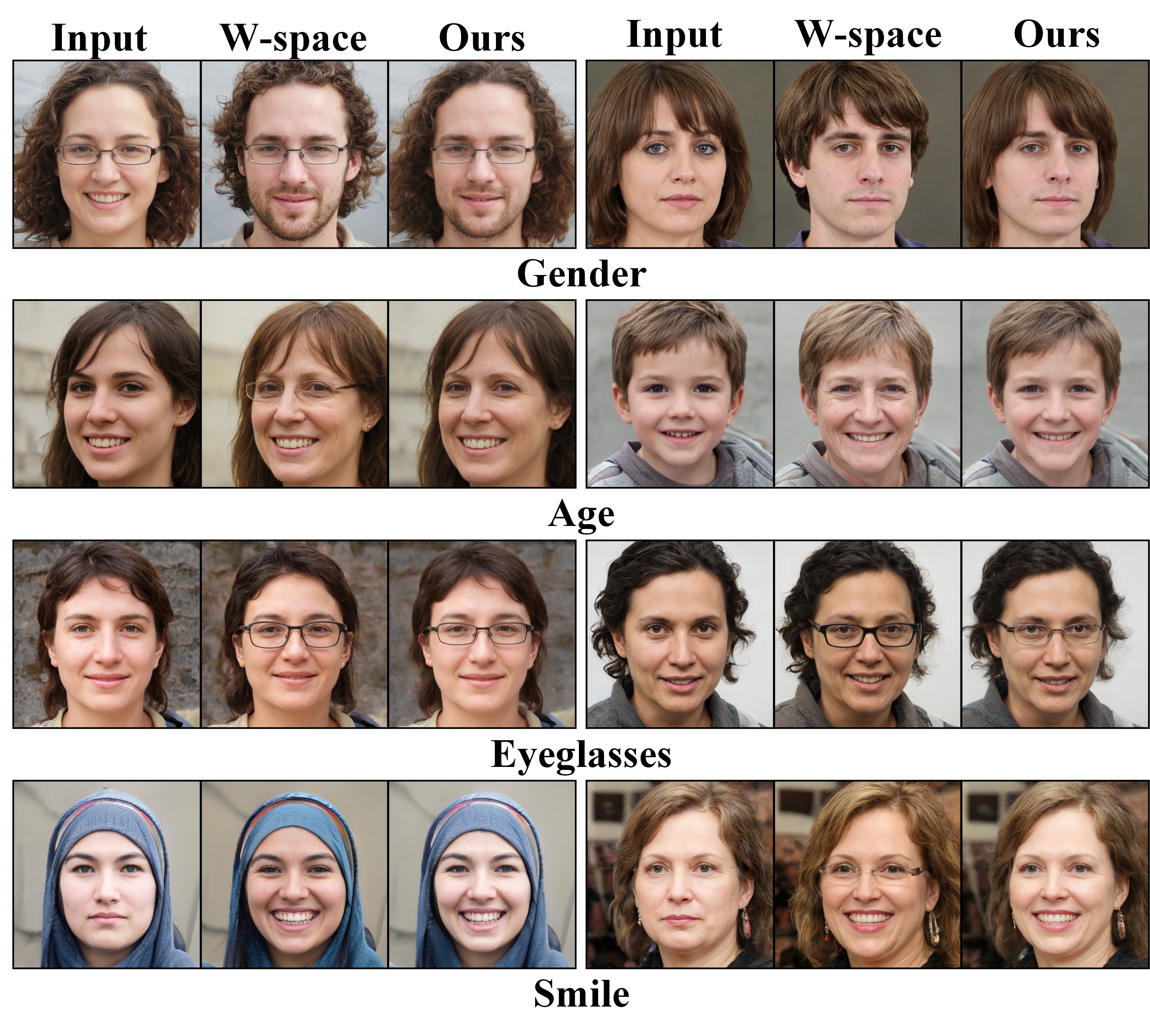}
\end{center}
\caption{Comparison between our method and editing in the $\mathcal{W}$-space. Our method still outperforms editing in $\mathcal{W}$-space by achieve more precise image translations.}
\label{fig:comparison_w}
\end{figure}

\begin{table*}[t]
\caption{Comparison of quantitative results measured by MSE, SSIM and ID. `EG' is the abbreviation for `Eyeglasses'. The metric `ID' is the confidence score of face verification obtained by the Face++ online analysis toolkit.}
\label{table:quantitative_comparison} 
\centering
\resizebox{1.95\columnwidth}{!}{%
\begin{tabular} {lcccc|cccc|cccc}
\toprule
       & \multicolumn{4}{c}{MSE ($\downarrow$)} &\multicolumn{4}{c}{SSIM ($\uparrow$)} & \multicolumn{4}{c}{ID ($\uparrow$)} \\
       \cmidrule{2-5}            \cmidrule{6-9}              \cmidrule{10-13}
Method & Gender & Age & EG & Smile & Gender & Age & EG & Smile & Gender & Age & EG & Smile  \\
\midrule
$\mathcal{Z}$-space  & 0.034 & 0.015 & 0.022 & 0.016 & 0.55 & 0.61 & 0.59 & 0.61 & 75.78 & 86.60 & 83.81 & 92.83 \\
$\mathcal{W}$-space  & 0.018 & 0.011 & 0.010 & 0.016 & 0.60 & 0.64 & 0.64 & 0.59 & 79.17 & 80.89 & 90.65 & 84.06 \\
InterFaceGAN  & 0.040  & 0.013 & 0.013 & 0.015 & 0.55 & 0.63 & 0.64 & 0.62 & 77.83 & \textbf{89.73} & 93.67 & 93.05 \\
Ours   & \textbf{0.015} & \textbf{0.008} & \textbf{0.004} & \textbf{0.003} & \textbf{0.61} & \textbf{0.66} & \textbf{0.72} & \textbf{0.71} & \textbf{80.47} & 89.37 & \textbf{94.99} & \textbf{95.21} \\
\bottomrule
\end{tabular}
}
\end{table*}

\begin{figure*}[t]
\begin{center}
\includegraphics[width=0.95\linewidth]{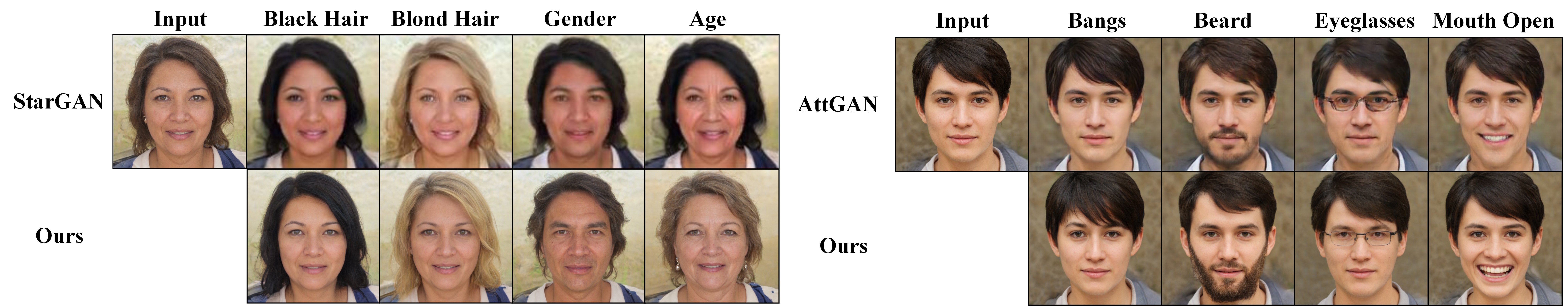}
\end{center}
\caption{Comparison between our method and approaches based on conditional GANs: StarGAN ($128\times 128$) and AttGAN ($256\times 256$).}
\label{fig:comparison_SOTA}
\end{figure*}

As the result shown in Figure~\ref{fig:comparison_z_interfacegan}, it is clear that our method could disentangle manipulations of the target facial component from other image content, and simultaneously render realistic translations.
On the contrary, editing in $\mathcal{Z}$-space clearly produces spatially entangled modifications to the input image.
Although InterFaceGAN largely solves the problem of semantic entanglement, as shown in Figure~\ref{fig:comparison_z_interfacegan} (b) and (d), the results still suffer from undesirable changes in terms of specific image content (\eg background texture and personal identity).
These unwanted image modifications could hardly be described by abstract attribute labels and thus could not be eliminated by subtracting the latent projection, unless supervision on the image level is imposed.

Notably, as shown in Figure~\ref{fig:comparison_z_interfacegan} (c), removing the translation along `Age' also eliminate the target attribute change `Eyeglasses'.
This is because eyeglasses are more likely to appear in images with aged faces, and thus these two visual concepts are heavily entangled in the latent space.
To solve this problem, most of previous studies on style-based generators resort to the intermediate latent space, \ie $\mathcal{W}$-space, for more disentangled representations.
Therefore, we also compare the performance of our method against editing in the $\mathcal{W}$-space.
Based on the results shown in Figure~\ref{fig:comparison_w}, it is clear that compared to editing in $\mathcal{W}$, our method could still achieve better spatial disentanglement in manipulating individual visual concept.

To provide a more comprehensive evaluation of our method, quantitative experiments are also conducted.
We adopt Mean Squared Error (MSE), Structural Similarity Index Measure (SSIM), and face verification score (ID) as metrics to measure how image content of the input is preserved in generation results at different levels.
According to results reported in Table~\ref{table:quantitative_comparison}, our method outperforms benchmark approaches almost in all cases, demonstrating the effectiveness of the proposed method in rendering precise translations of local regions.
This is desirable in various practical applications, such as portrait image editing, where the accurate manipulation of individual object is required.

\begin{table}[t]
\caption{Results of user study. For each question, the proportion ($\%$) of votes favoring our method is reported.}
\centering
\resizebox{0.9\columnwidth}{!}{%
\begin{tabular}{lrr}
\toprule
Metric   & vs. AttGAN & vs. StarGAN \\
\midrule
Attribute Translation  & $82.85$  & $78.08$  \\
Attribute Preservation & $98.10$  & $94.76$  \\
Image Quality          & $100.00$ & $99.53$ \\
\bottomrule
\end{tabular}
}
\label{table:user_study}
\end{table}

\subsubsection{GANs with Traditional Structures}
Since traditional GANs, \ie GAN models with successive convolutional layers instead of style injection, also integrate supervision on the consistency of image content, we also compare the performance of our method to these models for a comprehensive analysis.
Concretely, we consider StarGAN~\cite{choi2018stargan} and AttGAN~\cite{he2019attgan} as representatives for state-of-the-art FAE methods based on conditional GANs.
The officially released models are used for testing, and we resize input images to corresponding resolutions~\footnote{StarGAN: https://github.com/yunjey/stargan}~\footnote{AttGAN: github.com/elvisyjlin/AttGAN-PyTorch}.

According to results shown in Figure~\ref{fig:comparison_SOTA}, similar to traditional models which are designed to preserve image content irrelevant to the target translation, our method could also maintain the consistency between input and output images.
We also conduct user study to comprehensively compare the performance between our method and traditional cGANs.
45 subjects participate in the survey and each is presented by 110 samples.
More specifically, each subject is asked to rank the performance of these methods on three aspects,~\ie target editing, input preservation, and image quality.
According to the result shown in Table~\ref{table:user_study}, our method is preferred by the vast majority of users.
Notably, more than $94.76\%$ subjects think the proposed method outperforms previous approaches in preserving non-target image content.
This demonstrates the effectiveness of manipulating individual style code in achieving spatial disentanglement, since all these methods impose supervision on image-level consistency.

\section{Conclusion}
Due to the intrinsic complexity of style-based generators and the spatial invariance of AdaIN normalization, image translations obtained by recent studies are spatially entangled and thus it is undesirable in practical applications.
In this paper, we lift the dimension of feature space and propose to intervene the style code directly for the precise manipulation of individual feature map.
A lightweight and flexible optimization-based algorithm is proposed based on the in-depth observation of internal mechanism of style-based generators.
Both qualitative and quantitative results have demonstrated the ability of our method in achieving spatially disentangled translation the target object could be manipulated alone.
Our method could be easily generalized to tasks beyond facial attribute editing, and may possibly inspire the future design of generators to incorporate information of spatial layout of images.

{\small
\bibliographystyle{ieee_fullname}
\bibliography{egbib}
}

\end{document}